\documentclass[11pt]{article}

\usepackage{acl}
\usepackage{times}
\usepackage{latexsym}
\usepackage[T1]{fontenc}
\usepackage[utf8]{inputenc}
\usepackage{microtype}

\usepackage{graphicx}
\usepackage{booktabs}
\usepackage{multirow}
\usepackage{amsmath, amssymb}
\usepackage{xcolor}
\usepackage{hyperref}
\usepackage{xspace}
\usepackage{enumitem}

\microtypesetup{protrusion=true,expansion=true,final}
\newcommand{\climb}{CLIMB\xspace}

\title{Most Transformer Modifications Still Do Not Transfer at 1--3B:\\
       A 2020--2026 Update to Narang et al.\ (2021)\\
       with Downstream Evaluation and a Noise Floor}

\author{
  Yang Zhao$^{*}$, \quad
  Jiahao Lu$^{*}$, \quad
  Bin Huang, \quad
  Guhua Zhang$^{\dagger}$, \quad
  Jie Zhou$^{\dagger}$ \\[2pt]
  Tencent \\[2pt]
  {\normalfont\small
   yzyangzhao@tencent.com, \;
   ricardojhlu@tencent.com, \;
   lingyuhuang@tencent.com, \;
   drakezhang@tencent.com} \\[4pt]
  {\normalfont\small
   $^{*}$Equal contribution. \quad
   $^{\dagger}$Co-corresponding authors.}
}

\begin{document}
\maketitle

\begin{abstract}
\citet{narang2021transformer} evaluated 40+ Transformer
modifications at T5-base scale and concluded that most did not
transfer. Five years later, the typical working regime has moved
to 1--3B parameters, downstream evaluation has replaced pretraining
perplexity, and a substantially different catalogue of
modifications has emerged. We revisit their question by testing
20 post-2021 Transformer modifications at 1.2B and 3B under strict
iso-data, iso-compute, iso-recipe control, with a multi-seed
baseline noise floor and \climb-12 downstream evaluation as the
primary metric. The central finding reproduces theirs at this
curated set: most modifications do not transfer. Of the 20
modifications, only two clear Bonferroni correction at 1.2B; one
of those two further fails to train stably at 3B under the shared
recipe. We also find that the loss--downstream gap reported by
\citet{tay2023scaling} enlarges several-fold for attention-output
modifications: two significant failures converge to within 2--3\%
of baseline validation loss yet drop 6--16 \climb-points. We conclude
that noise-floor reporting, downstream evaluation, and cross-scale
stability testing are now prerequisites for architecture
comparisons at 1--3B.
\end{abstract}

\section{Introduction}
\label{sec:intro}

\citet{narang2021transformer} tested 40-plus Transformer modifications
on T5-base ($\sim$220M parameters) under a shared codebase and
training recipe, and reported that most did not transfer: most
reported gains either disappeared or fell within seed variance under
controlled comparison. The study became the standard reference for
whether a proposed architecture change is likely to survive
re-implementation.

Five years later, three conditions of that comparison have changed.
\emph{First, the scale has moved.} 1--3B parameters is now the
typical working regime for open small-LM releases and controlled
architecture comparisons, ten times the T5-base size.
\emph{Second, the metric has moved.} Pretraining perplexity, Narang
et al.'s primary metric, has since been shown to be an unreliable
predictor of downstream task accuracy across architectures and
scales~\citep{tay2023scaling,liu2023samepreloss,lourie2025scalingunreliable}.
\emph{Third, the catalogue has moved.} A new generation of
modifications---softmax
replacements~\citep{ramapuram2024sigmoid,nakanishi2025ssmax,zuhri2025softpick},
attention regularizers~\citep{henry2020query,leviathan2024selective,ye2024diff,zhou2025value},
residual variants~\citep{pagliardini2024denseformer,zhu2024hyper,touvron2021layerscale,kimi2026attnres},
and FFN activations~\citep{shazeer2020glu}---has been proposed since,
none of which appear in Narang et al.'s original 40+ set.

We revisit Narang et al.'s question under these three changes. We
build a controlled benchmark of 20 intra-Transformer modifications
proposed since their study and train each under a fixed 1.2B
(and separately 3B) Llama-style baseline configuration, sharing the
same mixed-domain English corpus, optimizer recipe, tokenizer, and
data shuffle (§\ref{sec:method}). Two further protocol updates: we establish a
baseline noise floor by running three independent seeds each of the
baseline and the strongest modification, and adopt the seed-level
distribution as the significance threshold (§\ref{sec:method-noise});
and we evaluate on \climb{}-12 downstream benchmarks rather than on
pretraining loss.\footnote{We adopt the 12-task downstream evaluation
protocol of \citet{diao2025climb}, which we refer to as
\climb{}-12. The corpus we pretrain on, CLIMB-Mix-400B, is
released alongside that protocol (Appendix~\ref{app:data});
specific tasks are listed in §\ref{sec:method-eval}.}

\paragraph{Main finding.}
Narang et al.'s central conclusion reproduces at 1.2B. Of the 20
methods, eight fall within the baseline bootstrap 95\%
interval ($|z|<2$); six are nominal degraders ($|z|>2$); six are
nominal improvers. Under Bonferroni at $m{=}19$ (one test per
non-baseline method), two improvers survive; five degraders remain
significant. A 3B
robustness check (13 attempted; ten complete, three diverge) adds a
second filter. The ten completing runs (1 baseline, 7 improver
candidates with positive \climb-avg delta, 2 failure references)
all preserve the sign of their 1.2B effect; ordering within the
improver band reshuffles (Spearman $\rho{=}{-}0.27$). One of the
two 1.2B Bonferroni survivors fails to train stably at 3B across
all three noise-floor seeds (42, 43, 44): each diverges to NaN in a narrow 2800--3500
step window with a reproducible Post-LN-on-FFN variance growth
signature~\citep{xiong2020layernorm} rather than seed noise. A
1.2B-significant gain therefore does not automatically imply 1--3B
deployability.

\paragraph{Downstream matters.}
The loss-downstream decoupling identified by \citet{tay2023scaling}
at architecture-family scale enlarges several times over in the
attention-modification setting at 1.2B. Sigmoid Attention's
validation loss is only 2.4\% above baseline and SSMax's only
3.0\%, yet their \climb{} accuracies drop by 16 and 6 percentage
points. A loss-only ranking would place both near baseline. The
residual-side failure AttnRes, whose loss rises by 13\%, has a
\climb-avg drop consistent with that loss gap
(Appendix~\ref{app:diagnostics}); the decoupling is concentrated
in attention-output modifications, not a general
``loss is unreliable'' claim.

\paragraph{Contributions.}
The contribution is the protocol update, not a new architecture:
\begin{itemize}[leftmargin=1.4em,topsep=2pt,itemsep=1pt,parsep=0pt]
  \item[(i)] A controlled benchmark of 20 post-2021 modifications
    under one shared recipe, with a 10-method 3B robustness check.
  \item[(ii)] A noise-floor protocol calibrated on three three-seed
    references (baseline, QK-Norm, Softpick), with
    $\sigma_\text{base}$ used as the conservative common threshold
    under Bonferroni / Holm / BH correction.
  \item[(iii)] A cross-hardware baseline reproduction (NVIDIA A100
    vs.\ primary accelerator) matching \climb-avg to
    $1\!\times\!10^{-4}$.
  \item[(iv)] Three recommendations (R5--R7,
    §\ref{sec:discussion}) extending Narang et al.'s original four:
    multi-seed noise floors, downstream evaluation for
    attention-output modifications, and cross-scale stability
    checks.
\end{itemize}

\section{Related Work}
\label{sec:related}

\subsection{Controlled comparative studies}
\label{sec:rel-controlled}

The direct ancestor of the present study is \citet{narang2021transformer},
who evaluated 40-plus modifications of the T5 encoder-decoder
architecture under a controlled recipe at T5-Base scale ($\sim$220M
parameters) and concluded that most proposed modifications did not
transfer. That paper established the standard methodological frame
for post-hoc audits of Transformer architecture claims---iso-codebase,
iso-recipe, single-seed pretraining-perplexity comparison---and is
the reference against which individual architecture improvements are
judged. We adopt the same spirit and update the protocol along three
axes that have since become binding: scale (1--3B rather than
T5-Base), metric (downstream rather than perplexity), and statistical
threshold (multi-seed noise floor with family-wise correction rather
than single-seed point estimates). The modification set is disjoint
from theirs by construction. \citet{tay2023scaling} compare ten
architecture \emph{families} from 15M to 40B parameters and document
a systematic gap between upstream pretraining perplexity and
downstream task performance; their observation motivates our metric
change and is the closest prior statement of the upstream--downstream
decoupling Appendix~\ref{app:diagnostics} analyzes. Their comparison
is at the level of architecture families; ours is at the level of
within-Transformer modifications, where the decoupling turns out to
be several times larger in magnitude for post-2024 attention-output
changes. \citet{liu2023samepreloss} and
\citet{lourie2025scalingunreliable} provide additional evidence that
pretraining loss is an unreliable downstream predictor in adjacent
settings.

\subsection{Modifications evaluated in this study}
\label{sec:rel-methods}

\paragraph{Soft modifications of softmax attention.}
QK-Norm~\citep{henry2020query,dehghani2023vit22b} applies an RMSNorm
to query and key projections before the dot product. Selective
Attention~\citep{leviathan2024selective} adds a learnable per-position
mask that down-weights distractor keys. Differential
Transformer~\citep{ye2024diff} computes two softmax distributions per
head and returns their difference. Value
Residual~\citep{zhou2025value} threads a shortcut from value
projections to subsequent layers. Softpick~\citep{zuhri2025softpick}
rectifies softmax,
$\text{softpick}(z)=\text{ReLU}(\text{softmax}(z)-1/n)$, producing
exact zeros for below-uniform weights.

\paragraph{Hard replacements of softmax.}
\citet{ramapuram2024sigmoid} replace softmax with element-wise
sigmoid, arguing the absence of cross-position normalization can
ease long-context scaling. SSMax~\citep{nakanishi2025ssmax} is a
scalable-softmax variant with a learned per-head temperature. Our
results (§\ref{sec:results}, Appendix~\ref{app:diagnostics}) indicate
that at 1.2B, removing softmax's normalization across positions is
not a productive modification on \climb{}-12.

\paragraph{Residual and inter-layer connectivity.}
DenseFormer~\citep{pagliardini2024denseformer} replaces the standard
residual with a depth-weighted average of all previous layers'
outputs. HyperConnections~\citep{zhu2024hyper} introduce
multi-channel inter-layer connectivity and have been used in
DeepSeek-V3~\citep{deepseekai2024v3} ($\sim$671B parameters).
LayerScale~\citep{touvron2021layerscale} adds a learnable
per-channel gate to the residual branch.
AttnRes~\citep{kimi2026attnres} replaces the identity residual
after attention with softmax attention over preceding layer outputs;
we adopt the pseudo-query variant
(Appendix~\ref{app:impl-attnres}).

\paragraph{Normalization and FFN.}
Sandwich Norm~\citep{ding2021cogview} wraps both Pre- and Post-LN
around each sublayer. HybridNorm~\citep{zhuo2025hybridnorm} mixes
attention-side Pre-LN with FFN-side Post-LN. Gated
Attention~\citep{qiu2025gated} adds a per-head sigmoid gate on the
attention output. \citet{shazeer2020glu} proposes GLU variants
including SwiGLU (the baseline used here) and GeGLU.
\citet{so2022primer} introduce Primer's non-gated ReLU$^{2}$
activation, since adopted by PaLM-540B.

\subsection{FLOPs accounting}
\label{sec:rel-flops}

The $6N$ approximation from Chinchilla scaling
laws~\citep{hoffmann2022chinchilla} does not distinguish attention's
$O(s^2 d)$ cost from the FFN's
$O(s\!\cdot\! d\!\cdot\! d_\text{inter})$ cost. We compute
operation-level per-step FLOPs (Appendix~\ref{app:flops}) so that
method-to-method compute differences can be read directly from
Table~\ref{tab:main-results}.

\section{Experimental Methodology}
\label{sec:method}

\begin{table*}[!t]
\centering
\small
\setlength{\tabcolsep}{6pt}
\renewcommand{\arraystretch}{1.15}
\begin{tabular}{@{}l l l@{}}
\toprule
\textbf{Category} & \textbf{Subcategory} & \textbf{Methods} \\
\midrule
Baseline      & ---                     & Baseline \\
\midrule
Attention     & QK-norm                 & QK-Norm \\
              & Learnable gate          & Selective Attention, Selective + QK-Norm \\
              & Softmax cap             & Softmax-Cap \\
              & Scaled softmax          & SSMax \\
              & Value shortcut          & Value-Residual \\
              & Sigmoid replacement     & Sigmoid Attention \\
              & Rectified softmax       & Softpick \\
              & Head subtraction        & Diff-Attn \\
              & Output gate             & Gated + QK-Norm \\
\midrule
FFN           & Gated (GeGLU)           & GeGLU \\
              & Non-gated ReLU$^{2}$    & ReLU$^{2}$ \\
              & GeGLU + QK-Norm stack   & QK-Norm + GeGLU \\
\midrule
Normalization & Sandwich                & Sandwich Norm \\
              & Attn Pre + FFN Post     & HybridNorm \\
\midrule
Residual      & Depth average           & DenseFormer \\
              & Learnable gate          & LayerScale \\
              & Multi-channel           & HyperConnections \\
              & Cross-layer pseudo-$Q$  & AttnRes \\
\bottomrule
\end{tabular}
\caption{The 20 Transformer modifications studied (1 baseline + 10
attention + 3 FFN + 2 normalization + 4 residual), grouped by the
module modified. The set covers the Attention, FFN, Normalization,
and Residual categories present in
\citet{narang2021transformer}'s original 40+ comparison, expanded
with methods proposed between 2020 and 2026. Within the Attention
column a soft/hard label indicates whether softmax remains the
primary mixing operation: QK-Norm, Selective Attention, SSMax,
Value-Residual and Softmax-Cap apply softmax to a (possibly
transformed) $QK^\top$ with at most an element-wise gate on the
result and are \emph{soft}; Sigmoid replaces softmax with a
different nonlinearity, Softpick rectifies softmax in a way that
breaks the row-sum, and Diff-Attn combines two softmaxes via
subtraction that can produce negative weights---these three are
\emph{hard}.}
\label{tab:methods}
\end{table*}

\subsection{Methods and taxonomy}
\label{sec:method-taxonomy}

We select 20 Transformer modifications proposed since
\citet{narang2021transformer}'s 2021 study, each claiming a gain
over a standard Llama-style baseline (or, for one combination
variant QK-Norm + GeGLU, representing a production stack of
standard components). The primary grouping
(Table~\ref{tab:methods}) is by the \emph{module} the modification
targets: attention, FFN, normalization, residual.

The 20-method set is not a census of the post-2021 catalogue, nor a
random sample. It is curated to cover the categories that
\citet{narang2021transformer}'s original 40+ set spanned (attention,
FFN, normalization, residual, softmax). Sandwich
Normalization~\citep{ding2021cogview}, Hybrid
Normalization~\citep{zhuo2025hybridnorm},
ReLU$^{2}$~\citep{so2022primer}, and Gated
Attention~\citep{qiu2025gated} were added to correct
under-representation relative to Narang et al.'s set, where
activation and normalization changes accounted for $\sim$30\% of the
tested modifications. Inclusion required (a) targeting a vanilla
decoder Transformer rather than a non-Transformer alternative
(state-space, linear attention), (b) public implementation or
sufficient detail to reproduce, and (c) at least one published
evaluation at $\geq$$0.5$B parameters. Claims of the form ``most
modifications do not transfer'' are scoped to this curated set.

Within the attention column, a secondary label
(\emph{soft}/\emph{hard}) records whether $\text{softmax}$ remains the
primary mixing operation: \emph{soft} if the attention block uses
$\text{softmax}$ of a (possibly transformed) $QK^\top$ as its mixing
weights, with at most element-wise scalar reweighting of the result
(QK-Norm, Selective Attention, SSMax, Value-Residual, Softmax-Cap);
\emph{hard} if it replaces softmax with a different nonlinearity
(Sigmoid Attention), rectifies the output to break row-stochasticity
(Softpick), or combines multiple softmaxes in a way that admits
negative weights (Differential Transformer).

\subsection{Training setup}
\label{sec:method-training}

The 1.2B baseline is a decoder-only Llama-2-style
Transformer~\citep{touvron2023llama}: 24 layers, hidden 2048, 32
attention heads (8 KV heads, GQA), SwiGLU FFN intermediate 5632,
RoPE, 65{,}664-token BPE vocabulary, context 1024. The 3B variant
(§\ref{sec:3b}) scales to 28 layers and hidden 3072 with all other
details identical. Method-specific hyperparameters (e.g.\ softmax
cap threshold) follow original papers; all other architectural
hyperparameters are held fixed. We pre-tokenize a mixed-domain
English corpus into 1024-token sequences; the same pack and
shuffle seed are used across every 1.2B run; 3B runs consume a
superset of the 1.2B shards (Appendix~\ref{app:data} reports the
exact 23.28B / 60.04B subset/superset splits). All 1.2B runs train
for 44{,}000 steps at $2^{20}$ tokens/step (23.07B training
tokens). Optimization: AdamW, lr $3\!\times\!10^{-4}$,
$\beta_1{=}0.9$, $\beta_2{=}0.95$, weight decay 0.1, gradient
clipping 1.0, 2{,}000 linear warm-up, cosine decay to 10\%, bf16
with FP32 master weights. No per-method tuning. Each 1.2B run uses
one node of eight AI accelerators ($\sim$16h); each 3B run
uses two RDMA-connected nodes ($\sim$3 days). Hardware vendor
anonymized for review.

\paragraph{Numerical-stability disclosure.}
At 3B, three methods (HybridNorm, SSMax, HyperConnections) diverge
under the shared recipe; the baseline and 9 other 3B-attempted
modifications train to completion. Each divergence has documented
theoretical fragility (Post-LN deep-layer variance
growth~\citep{xiong2020layernorm}, softmax-replacement logit
blowup~\citep{ramapuram2024sigmoid,nakanishi2025ssmax}, multi-channel EMA
residual drift~\citep{zhu2024hyper}) independent of accelerator;
observed grad-norm signatures are mechanism-aligned
(§\ref{sec:3b-failures}). The possibility that a tuned recipe could
stabilize any of the three is discussed in Limitations.

\subsection{Noise-floor protocol}
\label{sec:method-noise}

We establish a noise floor by running three independent seeds
(42, 43, 44) each of three reference configurations: the baseline,
the strongest \emph{soft} modification (QK-Norm), and the strongest
\emph{hard} modification (Softpick). Each seed varies both
initialization and data-shuffle RNG. Three-seed empirical standard
deviations of \climb-avg (unbiased):
$\sigma_\text{baseline}{=}0.00208$, $\sigma_\text{QK-Norm}{=}0.00146$,
$\sigma_\text{Softpick}{=}0.00133$. The baseline calibration is the
most conservative; we use $\sigma_\text{baseline}$ operationally so
improver $z$-scores are if anything understated. The primary
significance test is a seed-only bootstrap with $N{=}10{,}000$
resamples drawing three baseline seeds with replacement: the
QK-Norm and Softpick 95\% intervals are disjoint from baseline's
($\Pr(\text{method}{\le}\text{base}){=}0$ in the paired bootstrap).
For a single-seed method with \climb-avg $x$, we report
$z=(x-\hat\mu_\text{base})/\hat\sigma_\text{base}$ and flag $|z|>2$
as significant. Per-method cross-checks not inflated by catalogue
size: 3-seed Welch's $t$ gives $t{=}{+}4.39$, $p{=}0.015$ for
QK-Norm and $t{=}{+}6.36$, $p{=}0.0053$ for Softpick. A per-task
Welch + Stouffer combination is reported in
Appendix~\ref{sec:diag-stouffer}.

\subsection{Downstream evaluation suite}
\label{sec:method-eval}

We evaluate every model on \climb{}-12: PIQA~\citep{bisk2020piqa},
ARC-Challenge and ARC-Easy~\citep{clark2018arc},
HellaSwag~\citep{zellers2019hellaswag},
WinoGrande~\citep{sakaguchi2020winogrande},
SocialIQA~\citep{sap2019socialiqa}, MMLU~\citep{hendrycks2021mmlu},
OpenBookQA~\citep{mihaylov2018openbookqa}, BoolQ~\citep{clark2019boolq},
RACE~\citep{lai2017race}, LAMBADA~\citep{paperno2016lambada}, and
TruthfulQA-MC2~\citep{lin2022truthfulqa}, using the default few-shot
configuration in
lm-evaluation-harness~\citep{eval-harness}; \climb-avg is
the unweighted macro-average. Downstream rather than pretraining
loss is the primary metric because nearly-identical validation
losses can correspond to 6--16-point \climb{} differences
(Appendix~\ref{app:diagnostics}). Optimizer, schedule, tokenizer,
sequence length, batch size, RoPE base, initialization, and data
shuffle are fixed across all 20 methods. Per-step FLOPs are computed
by operation-level counting (Appendix~\ref{app:flops}); all 20
methods lie within $\pm 0.13\%$ of baseline.

\section{Results at 1.2B}
\label{sec:results}

\subsection{Overview}
\label{sec:results-overview}

\begin{table*}[!t]
\centering
\small
\setlength{\tabcolsep}{6pt}
\begin{tabular}{@{}lrrrr@{}}
\toprule
\textbf{Method} &
\textbf{\climb-avg} & \textbf{$\Delta$} &
\textbf{$z$} & \textbf{$p_\text{Bonf}$} \\
\midrule
Softpick             & 0.4922 & $+0.009$ & $+4.47$  & $1.5\!\times\!10^{-4}$ \\
HybridNorm$^{\ddagger}$ & 0.4896 & $+0.007$ & $+3.19$ & $0.027$ \\
QK-Norm              & 0.4882 & $+0.005$ & $+2.51$  & $0.23$ \\
Sandwich Norm        & 0.4880 & $+0.005$ & $+2.45$  & $0.27$ \\
ReLU$^{2}$           & 0.4878 & $+0.005$ & $+2.36$  & $0.35$ \\
Selective Attention  & 0.4874 & $+0.004$ & $+2.15$  & $0.61$ \\
QK-Norm + GeGLU      & 0.4863 & $+0.003$ & $+1.60$  & $1.00$ \\
Gated + QK-Norm      & 0.4855 & $+0.003$ & $+1.23$  & $1.00$ \\
Softmax-Cap          & 0.4854 & $+0.002$ & $+1.18$  & $1.00$ \\
Selective + QK-Norm  & 0.4852 & $+0.002$ & $+1.11$  & $1.00$ \\
Value-Residual       & 0.4846 & $+0.002$ & $+0.81$  & $1.00$ \\
GeGLU                & 0.4834 & $+0.001$ & $+0.20$  & $1.00$ \\
Baseline             & 0.4829 & $\phantom{+}0.000$ & $\phantom{+}0.00$ & --- \\
DenseFormer          & 0.4809 & $-0.002$ & $-0.97$  & $1.00$ \\
Diff-Attn            & 0.4779 & $-0.005$ & $-2.45$  & $0.27$ \\
LayerScale           & 0.4738 & $-0.009$ & $-4.42$  & $1.8\!\times\!10^{-4}$ \\
HyperConnections     & 0.4626 & $-0.020$ & $-9.79$  & $2.6\!\times\!10^{-21}$ \\
AttnRes              & 0.4218 & $-0.061$ & $-29.47$ & ${<}10^{-7}$ \\
SSMax                & 0.4208 & $-0.062$ & $-29.94$ & ${<}10^{-7}$ \\
Sigmoid Attention    & 0.3217 & $-0.161$ & $-77.68$ & ${<}10^{-7}$ \\
\bottomrule
\end{tabular}
\caption{Main results at 1.2B, sorted by \climb-avg. The Baseline
row reports the three-seed mean $\bar{x}_\text{base}{=}0.4829$ used
as the reference for $\Delta$ and $z$ throughout: $\Delta = x -
\bar{x}_\text{base}$, $z = \Delta / \sigma_\text{base}$, with
$\sigma_\text{base}{=}0.00208$ (unbiased 3-seed empirical std;
§\ref{sec:method-noise}). $p_\text{Bonf}$: Bonferroni at $m{=}19$
(one test per non-baseline method; the Baseline row is the $z{=}0$
reference, not a tested hypothesis).
$|z|{>}20$ should be read as ``substantially below baseline noise.''
$^{\ddagger}$~HybridNorm passes Bonferroni at 1.2B but diverges at 3B.}
\label{tab:main-results}
\end{table*}

Table~\ref{tab:main-results} reports each method's \climb-avg, delta
relative to baseline, bootstrap $z$-score, and per-step FLOPs. Most
modifications in this curated set do not transfer, consistent with
\citet{narang2021transformer} at T5-base. Of the 20 methods, eight
land within the baseline bootstrap 95\% interval ($|z|<2$); six
significantly degrade baseline; six nominally improve at $|z|>2$,
$p<0.05$---Softpick~\citep{zuhri2025softpick} ($z{=}{+}4.47$),
HybridNorm~\citep{zhuo2025hybridnorm} ($z{=}{+}3.19$),
QK-Norm~\citep{henry2020query} ($z{=}{+}2.51$),
Sandwich Norm~\citep{ding2021cogview} ($z{=}{+}2.45$),
ReLU$^{2}$~\citep{so2022primer} ($z{=}{+}2.36$), and Selective
Attention~\citep{leviathan2024selective} ($z{=}{+}2.15$). Under
Bonferroni at $\alpha{=}0.05/19$ (one test per non-baseline method), Softpick
($p_\text{Bonf}\!\approx\!1.5\!\times\!10^{-4}$) and HybridNorm
($\approx\!0.027$) survive; the other four reduce to suggestive
evidence. The five large-effect degradations remain significant
after correction; Diff-Attn does not. Per-method 3-seed Welch's $t$
gives QK-Norm $t{=}{+}4.39$, $p{=}0.015$ and Softpick $t{=}{+}6.36$,
$p{=}0.0053$. Holm--Bonferroni preserves the same 7 methods;
Benjamini--Hochberg at FDR$=0.05$ adds QK-Norm, Sandwich Norm,
ReLU$^{2}$, and Diff-Attn. The qualitative conclusion is the same
under all three procedures. Of the two 1.2B Bonferroni survivors,
only one trains to completion at 3B under the shared recipe; the
other diverges at step 3500 with a Post-LN-compatible signature
(§\ref{sec:3b-failures}).

\subsection{Attention modifications span the largest effect range}
\label{sec:results-attention}

The ten attention modifications (Table~\ref{tab:methods}) span
\climb-avg from $+0.009$ (Softpick) to $-0.161$ (Sigmoid Attention)
---roughly eight times the baseline bootstrap width. Six of the eight
non-combination attention modifications are statistically
distinguishable from baseline, three improving and five degrading; only
Value-Residual and Softmax-Cap fall inside the noise band. Modifying
the attention normalization is therefore a high-variance
intervention: small perturbations such as QK-Norm shift the result by
a few $\sigma_\text{base}$ in the positive direction, while
softmax-replacement variants (Sigmoid, SSMax) overshoot the negative
direction by two orders of magnitude over the noise floor. The
soft/hard label of Table~\ref{tab:methods} does not predict outcome
at the level of individual methods: the single Bonferroni-surviving
improvement (Softpick) is hard, the largest improvers besides
Softpick are soft, and the largest failure (Sigmoid) is hard. The
label is a structural taxonomy, not a predictor.

\subsection{Residual and FFN modifications}
\label{sec:results-residual}

None of the four residual-connection modifications improves over
baseline; three are significantly worse ($z{=}{-}4.4$, ${-}9.8$,
${-}29.5$); DenseFormer is statistically indistinguishable
($z{=}{-}0.97$). On FFN: GeGLU ($z{=}{+}0.20$) is indistinguishable
from the SwiGLU baseline; ReLU$^{2}$~\citep{so2022primer}, evaluated
at iso-parameter count, reaches $z{=}{+}2.36$ but not Bonferroni;
QK-Norm + GeGLU (the production stack of Llama-3, Gemma-2) reaches
the top of the 3B rank at iso-token (§\ref{sec:3b-top5}) despite an
uninformative 1.2B $z{=}{+}1.60$.

\subsection{Per-method evidence for Softpick}
\label{sec:results-softpick}

Of the two 1.2B Bonferroni survivors,
Softpick~\citep{zuhri2025softpick} also trains stably at 3B
(§\ref{sec:3b}). Softpick is a one-line modification of softmax,
$\text{softpick}(z)=\text{ReLU}\!\left(\text{softmax}(z)-1/n\right)$,
producing exact zeros for below-uniform attention weights. The
original paper reports it eliminates attention sinks and reduces
massive activations at scales up to 340M, evaluated primarily on
pretraining loss. The 1.2B and 3B controlled replication here uses
the same protocol as the other 19 methods, with full \climb-12
downstream evaluation and per-method multi-seed evidence. Three
Softpick seeds yield \climb-avgs $0.4922$, $0.4905$, $0.4931$
(mean $0.4919$, $\sigma{=}0.00133$); three baseline seeds yield
$0.4820$, $0.4815$, $0.4853$ (mean $0.4829$, $\sigma{=}0.00208$).
The 3-seed Welch's $t$-test ($t{=}{+}6.36$, $p{=}0.0053$) is a
per-method test that does not inflate with catalogue size. Under
the operational criterion of §\ref{sec:method-taxonomy}, Softpick
is \emph{hard}: ReLU rectification zeroes weights below $1/n$, so
the row no longer sums to one. The narrative ``replacing softmax
hurts'' has a counter-example introduced by
\citeauthor{zuhri2025softpick} and replicated here. What separates
this from the two significant hard failures (Sigmoid, Diff-Attn)
is that rectified softmax preserves \emph{peaked} attention:
weights surviving rectification are exactly the softmax weights
above uniform, so a single-key lookup pattern (e.g.\ LAMBADA) is
structurally supported (Appendix~\ref{app:diagnostics}).

\subsection{Most modifications are indistinguishable from baseline}
\label{sec:results-noise}

Eight of the 20 methods land inside the baseline bootstrap 95\%
interval ($|z|<2$): QK-Norm + GeGLU, Gated + QK-Norm, Softmax-Cap,
Selective + QK-Norm, Value-Residual, GeGLU, the baseline reference
itself ($z{=}0$ by construction), and DenseFormer.
\citeauthor{narang2021transformer}
report approximately five of 40+ modifications yielded a meaningful
T5-base improvement; our 20-method 1.2B sample (6 nominal improvers,
2 Bonferroni survivors, 1 surviving the 3B filter) is consistent in
order of magnitude. We do not claim our 20-method set is random or
exhaustive; the headline ``most do not transfer'' is scoped to this
curated set (§\ref{sec:method-taxonomy}).

\section{Robustness at 3B}
\label{sec:3b}

\begin{table*}[!t]
\centering
\small
\setlength{\tabcolsep}{6pt}
\begin{tabular}{@{}lrrrr@{}}
\toprule
\textbf{Method} & \textbf{1.2B} & \textbf{3B} &
\textbf{$\Delta$ 3B} & \textbf{rank} \\
\midrule
QK-Norm + GeGLU      & 0.4863 & \textbf{0.5158} & $+0.0169$            & 7$\to$1 \\
QK-Norm             & 0.4882 & 0.5129          & $+0.0139$            & 3$\to$2 \\
Selective + QK-Norm  & 0.4852 & 0.5108          & $+0.0118$            & 10$\to$3 \\
Selective Attention    & 0.4874 & 0.5097          & $+0.0107$            & 6$\to$4 \\
Sandwich Norm     & 0.4880 & 0.5088          & $+0.0099$            & 4$\to$5 \\
Softpick           & \textbf{0.4922} & 0.5084 & $+0.0095$            & 1$\to$6 \\
ReLU$^{2}$      & 0.4878 & 0.5083          & $+0.0093$            & 5$\to$7 \\
Baseline           & 0.4829 & 0.4989          & $\phantom{+}0.0000$ & 13$\to$8 \\
AttnRes            & 0.4218 & 0.4322          & $-0.0667$            & 19$\to$9 \\
Sigmoid Attention      & 0.3217 & 0.3187          & $-0.1803$            & 20$\to$10 \\
\bottomrule
\end{tabular}
\caption{3B \climb-avg at the iso-token milestone
($\sim$23B training tokens) alongside the 1.2B value from
Table~\ref{tab:main-results}. The 1.2B Baseline value is the
three-seed mean ($\bar{x}_\text{base}{=}0.4829$); the 3B Baseline
is single-seed. $\Delta$~3B is each method's \climb-avg minus the
3B single-seed baseline ($0.4989$); we have not run multi-seed 3B
calibration. ``rank'' is each method's position
(20 at 1.2B; 10 completed at 3B).}
\label{tab:3b}
\end{table*}

\subsection{Setup}
\label{sec:3b-setup}

Deployability claims need the harder question of whether 1.2B gains
\emph{transfer} to the next scale. We re-train a subset of the 1.2B
methods at 3B under the same iso-recipe (28 layers, hidden 3072,
all other details from §\ref{sec:method-training}). Each 3B run
trains to an iso-token milestone of $\sim$23B tokens
($44{,}000\!\times\!2^{20}$). No per-method retuning; a method that
diverges at 3B is reported as a scale-up failure, not discarded. A
full 20-method 3B sweep is approximately $2.5\times$ the 1.2B
budget; we instead train 13 configurations: the 3B baseline; seven
improver candidates (the five 1.2B nominal improvers other than
HybridNorm, plus two sub-Bonferroni catalogue methods QK-Norm + GeGLU
and Selective + QK-Norm); two 1.2B significant failures (Sigmoid
Attention, AttnRes); and three methods that diverge at 3B
(HybridNorm, SSMax, HyperConnections; §\ref{sec:3b-failures}).
HybridNorm passes 1.2B Bonferroni but is grouped with the 3B
divergences. Ten configurations complete; three diverge.

\subsection{Improver candidates transfer at 3B; rank reshuffles}
\label{sec:3b-top5}

The seven 1.2B improver candidates that train to completion at 3B
all maintain a positive \climb-avg delta against a 3B baseline
trained under the identical recipe (Table~\ref{tab:3b}). Deltas
range from $+0.009$ (ReLU$^{2}$) to $+0.017$ (QK-Norm + GeGLU). The two
1.2B-significant failures retain large negative deltas: Sigmoid
Attention drops by $-0.180$ at 3B (vs.\ $-0.161$ at 1.2B); AttnRes
drops by $-0.067$ (vs.\ $-0.061$). \emph{The sign of every 1.2B
signal is preserved at 3B}: 7/7 improvers stay positive, 2/2
failures stay negative. Relative ordering reshuffles substantially:
QK-Norm + GeGLU moves from 1.2B rank 7 ($z{=}{+}1.60$, not
Bonferroni-significant) to 3B rank 1; Selective + QK-Norm from
rank 10 to rank 3; Softpick, the 1.2B Bonferroni survivor that
completes 3B, slips to rank 6. Spearman $\rho{=}{-}0.27$ on the
seven improvers indicates 1.2B improver \emph{rank} is a weak
predictor of 3B improver rank; with $n{=}7$ and single-seed
measurements per method, this $\rho$ is far from significant
($p{\approx}0.56$ two-sided) and is reported as a directional
signal, not a precise estimate. The transferable signal is in the
\emph{sign} of the 1.2B effect, not fine ordering: 1.2B can
reliably filter \emph{whether} a modification helps but is too
noisy to rank \emph{which} of a set of in-noise improvers will
help most at the next scale.

\subsection{Three methods diverge at 3B; divergences are mechanism-aligned}
\label{sec:3b-failures}

Three methods that passed the 1.2B noise floor failed to train at
3B (Table~\ref{tab:3b-failures} in Appendix~\ref{app:diagnostics}
records when each diverged; Figure~\ref{fig:hybrid-spike} in
Appendix~\ref{app:diagnostics} shows the HybridNorm grad-norm
trajectory).

\paragraph{HybridNorm.}
HybridNorm~\citep{zhuo2025hybridnorm} combines Pre-LN attention
with Post-LN FFN. At 1.2B it is Bonferroni-significant
($z{=}{+}3.19$, $p_\text{Bonf}{=}0.027$). To rule out seed-specific
anomalies we retrained at 3B under all three noise-floor seeds
(42, 43, 44); all three diverge in a narrow 2800--3500 step window
shortly after the cosine peak (step $\sim$2000), with pre-NaN
grad-norm in the $0.18$--$0.19$ range identical to baseline's
stable trajectory. This is a known Post-LN deep-decoder
instability~\citep{xiong2020layernorm}: Post-LN amplifies activation
variance with depth, and the 28-layer 3B stack under peak learning
rate exceeds the threshold the 24-layer 1.2B stack retains. Three
seeds failing in the same window is mechanism, not noise.

\paragraph{SSMax and HyperConnections.}
Scalable-softmax~\citep{nakanishi2025ssmax} multiplies pre-softmax
logits by $s\cdot\log n$ with learnable $s$; at 3B the grad-norm
grows monotonically from $27$ at step $\sim$$4000$ to $1578$ at
step $5600$, then NaN at $5800$---the softmax-replacement
instability documented in \citet{ramapuram2024sigmoid}.
HyperConnections~\citep{zhu2024hyper} maintain an EMA state across
layers; our output-side reimplementation
(Appendix~\ref{app:impl-hyper}) is stable at 1.2B
($z_{1.2\mathrm{B}}{=}{-}9.79$, finite) but at 3B shows sustained
grad-norm inflation from step $\sim$15000, NaN at $18300$.
\citeauthor{zhu2024hyper}'s positive report is at 671B MoE with
full input-side lane mixing, which the shared residual interface
in our codebase does not expose.

\paragraph{Not hardware artefacts.}
Ten of thirteen 3B-attempted methods complete; the three that do not
each exhibit a mechanism-aligned signature (Post-LN variance growth,
softmax-replacement logit blowup, multi-lane EMA drift) independent
of the accelerator. An A100 cross-hardware reproduction
(Appendix~\ref{app:a100}) confirms this independently.

\subsection{3B as a second filter on top of 1.2B Bonferroni}
\label{sec:3b-filter}

Two methods pass Bonferroni at 1.2B; only one trains to completion
at 3B with a positive delta. The 3B robustness experiment therefore
removes one of the two 1.2B Bonferroni survivors. Reporting
implication: any sub-billion-scale improvement proposed for 1--3B
deployment should include either a 3B stability check under the
submitter's shared recipe, or an explicit disclosure of the scale
at which the gain was measured.

\section{Discussion}
\label{sec:discussion}

\paragraph{Why don't most modifications transfer?}
\citet{narang2021transformer} enumerate candidate explanations for
their negative finding. \emph{(1) Idiosyncratic codebase.} Unlikely:
the baseline is the community-standard decoder (RoPE, GQA, SwiGLU,
RMSNorm) of Llama-2/3, Gemma-2, Qwen, Mistral. \emph{(2) Insufficient
per-method tuning.} \citeauthor{narang2021transformer} tested this
as a UT case study and found 23 of 25 sweeps still under-performed;
we hold the recipe fixed because per-method tuning reintroduces the
confound we are trying to remove. The QK-Norm and Softpick 3-seed
results show that modifications that do transfer do so under the
shared recipe. \emph{(3) 1.2B is below scale.} The 3B experiment
partially tests this: positive 1.2B deltas are preserved at 3B;
methods whose authors report benefits at much larger scale
(HyperConnections at 671B MoE, AttnRes at large-MoE) do not
reproduce here. \emph{(4) Most post-2021 modifications genuinely do
not transfer at 1--3B.} The 1.2B-to-3B filter removes one of two
Bonferroni survivors; six of 20 significantly \emph{degrade}
baseline, and the degradations are mechanism-aligned (softmax
replacement without row-normalization, Post-LN variance growth, EMA
drift), suggesting non-transfer is a systematic property of
particular interventions rather than random noise.

\paragraph{Three further recommendations.}
\citet{narang2021transformer} close with four recommendations (try
the modification in multiple codebases; apply it to diverse
downstream tasks; keep hyperparameters fixed; report mean and
standard deviation across trials), all of which still apply at
1--3B. We add three the post-2021 evidence makes binding.

\paragraph{R5. Multi-seed baseline noise floor.}
At 1.2B the three-seed $\sigma$ on \climb-avg is $\approx\!0.0021$;
any claimed gain below $\sim\!0.005$ is within seed variance, and
eight of 20 modifications fall here. Reporting single-seed numbers
without a baseline distribution overstates the count of useful
methods.

\paragraph{R6. Downstream evaluation for attention-output modifications.}
\citet{tay2023scaling}'s loss--downstream decoupling at
architecture-family scale extends with several times the magnitude
here. Two significant negatives (Sigmoid Attention, SSMax) reach
within 2--3\% of baseline validation loss while dropping 6--16
\climb-points (Table~\ref{tab:main-results} together with
Table~\ref{tab:loss-vs-climb} in Appendix~\ref{app:diagnostics}); a
loss-only ranking would miss them. Residual-side failures (AttnRes,
LayerScale) have loss increases consistent with their \climb{}
drops; the decoupling is concentrated in attention output.

\paragraph{R7. Cross-scale stability check under the shared recipe.}
With 19 modifications tested against the same baseline, the
family-wise false-positive rate at $\alpha{=}0.05$ is $\sim$62\%;
Bonferroni leaves two survivors, of which one fails to train
stably at 3B. A
Bonferroni-surviving 1.2B improvement is therefore not sufficient
evidence of 1--3B deployability. A per-method 3-seed Welch's
$t$-test additionally supplies single-method evidence not inflated
by catalogue size.

\paragraph{Closing.}
The protocol is the contribution. Future 1--3B architecture studies
need three prerequisites: multi-seed baselines, downstream
evaluation, and cross-scale stability checks.

\bibliography{bib/refs}

\section*{Limitations}
\label{sec:limitations}

\paragraph{Scale ceiling.}
A method we report as diverging at 3B may become stable under a
retuned recipe; our claim is that it does not train under the
\emph{shared} recipe, the strong condition for reproducing the
``single protocol, many methods'' comparison of Narang et al.
Some 1.2B failures may become beneficial at scales we do not reach
(\citealt{zhu2024hyper} at 671B MoE; \citealt{kimi2026attnres} in
a large-MoE setting). The scope of our claims is ``at 1.2--3B on
\climb-12 under a shared recipe.''

\paragraph{Hardware generality.}
The main 1.2B and 3B comparisons run on a single brand of AI
accelerator in bf16 (vendor anonymized for double-blind review;
to be disclosed in the camera-ready version). To test for
hardware-specific numerical sensitivity we re-trained the 1.2B
baseline end-to-end on NVIDIA A100 under bit-identical model code,
recipe, data shuffle, and 23B-token budget. The two baselines
reach \climb-avg $0.4820$ (primary) vs.\ $0.4821$ (A100), a
difference of $1\!\times\!10^{-4}$, with per-task deltas within
$\pm 0.025$ (Appendix~\ref{app:a100}). This makes a blanket
infrastructure explanation for the three 3B divergences unlikely:
the baseline reproduces to $10^{-4}$ on A100 under the same
recipe, while the divergences each have documented theoretical
fragility independent of accelerator. We did not re-run the three
3B divergence cases on A100, so we cannot fully rule out an
accelerator-specific contribution to those particular failures; the
mechanism-aligned grad-norm signatures (§\ref{sec:3b-failures}) are
the strongest evidence we offer against that interpretation.

\paragraph{Single recipe, single corpus.}
Optimizer, schedule, warmup, initialization, and weight decay are
held fixed across all 20 methods, on a single mixed-domain English
corpus (23B training tokens per run; Appendix~\ref{app:data}). A
method whose original paper recommends a different recipe, or whose
benefits are tied to multilingual or code data, is disadvantaged
here. Per-method tuning would reintroduce the confound we removed;
the choice narrows the reach of the claims.

\paragraph{Noise-floor generalization.}
$\sigma$ is estimated from three calibration sets (baseline,
QK-Norm, Softpick): $0.00208$, $0.00146$, $0.00133$; we
use $\sigma_\text{baseline}$ as the most conservative. For
$|z|{>}5$ this is robust by inspection; for borderline methods
(Diff-Attn $-2.45$, Selective Attention $+2.15$,
ReLU$^{2}$ $+2.36$, Sandwich Norm $+2.45$),
the method's own $\sigma$ could change the call. Bootstrap CIs at
$n{=}3$ under-cover relative to
nominal~\citep{efron1987bootstrap}; the 95\% interval is a nominal
threshold rather than calibrated coverage.

\paragraph{Single-seed coverage at the borderline.}
Per-method 3-seed evidence exists only for QK-Norm and Softpick (the
two configurations selected as noise-floor anchors). The other four
nominal improvers above $|z|{>}2$ (HybridNorm $z{=}{+}3.19$,
Sandwich Norm $+2.45$, ReLU$^{2}$ $+2.36$, Selective Attention
$+2.15$) are evaluated under a single seed each, with significance
assessed against the baseline 3-seed bootstrap distribution. A 3-seed
Welch's $t$ on each would be a stricter test that does not inflate
with catalogue size; the compute cost (8 additional 1.2B runs) was
beyond this study's budget. The Bonferroni and Holm--Bonferroni
corrections we apply are conservative substitutes that already rule
all four as ``not significant'' after correction; the limitation is
that the per-method-evidence framing of R5 is met fully only for our
two anchor methods.

\paragraph{3B noise floor is single-seed except at the failure point.}
The 3B \climb{}-avg numbers in Table~\ref{tab:3b} use one seed per
method except for HybridNorm, which we re-ran under all three
noise-floor seeds (42, 43, 44) specifically to characterize its
divergence (§\ref{sec:3b-failures}). The 3B baseline is therefore
single-seed, so a 3B $\Delta$ within the 1.2B baseline noise band
($\sim$$0.0021$ \climb-avg) cannot be distinguished from baseline-
specific seed variance at 3B. Our 3B claims are restricted to (i)
the \emph{sign} of each method's $\Delta$ relative to the 3B
baseline (every 3B-completing improver shows $\Delta{>}0$ and both
3B-completing failures show $\Delta{<}0$ at sufficient magnitude
that single-seed seed variance does not flip the sign), and (ii)
\emph{stability outcomes} (which methods diverge under the shared
recipe). We do not draw fine rank conclusions among the 3B-completing
improvers from single-seed data; the Spearman $\rho{=}{-}0.27$
result in §\ref{sec:3b-top5} is reported as a coarse cross-scale
rank-instability signal, not an estimate. A multi-seed 3B
calibration would refine these claims and is left to future work.

\section*{Reproducibility}
\label{sec:reproducibility}

We aim to make the protocol of this paper reproducible end-to-end.

\paragraph{Code.}
All 20 method implementations described in Appendix~\ref{app:impl}
are written under a single shared training graph; only the module
under study differs between runs. Source code and configuration
files for every 1.2B and 3B run will be released at the
camera-ready stage. During review the implementations are
described in Appendix~\ref{app:impl} at sufficient detail to
re-implement from scratch (per-method equations, initialization
values, and any deviation from the original paper).

\paragraph{Training recipe.}
Optimizer (AdamW, lr $3\!\times\!10^{-4}$, $\beta_1{=}0.9$,
$\beta_2{=}0.95$, weight decay 0.1), gradient clipping (1.0),
warmup (2{,}000 linear), schedule (cosine to 10\% over the full
horizon), and precision (bf16 with FP32 master weights) are held
fixed across all 20 methods. The same configuration runs at both
1.2B and 3B (only the model dimensions differ:
\{24~layers, hidden~2048\} versus \{28~layers, hidden~3072\}).

\paragraph{Data and evaluation.}
The mixed-domain English pretraining corpus is pre-tokenized once,
and the same pack-and-shuffle order is used across every 1.2B run;
the 3B runs consume a superset of the same corpus under the same
shuffle construction (Appendix~\ref{app:data}; the 1.2B subset is
23.28B tokens, the 3B superset 60.04B tokens, with each individual
training run consuming 23B tokens at the iso-token milestone of
44{,}000 steps). All evaluation uses the
lm-evaluation-harness~\citep{eval-harness} default
few-shot configuration on the 12 \climb{} tasks listed in
§\ref{sec:method-eval}; \climb-avg is the unweighted macro-average.

\paragraph{Seeds and significance.}
Three independent seeds (42, 43, 44) are used for the baseline,
QK-Norm, and Softpick noise-floor calibrations. Each seed varies
both initialization and data-shuffle RNG. The bootstrap that
produces the $z$-scores in Table~\ref{tab:main-results} uses
$N{=}10{,}000$ resamples with replacement from the 3-seed baseline
distribution; the random-state seed for this bootstrap is fixed at
$0$ for reproducibility.

\paragraph{Hardware.}
The main 1.2B and 3B runs use eight-way data-parallel nodes of a
single brand of AI accelerator (vendor anonymized for
review, to be disclosed in the camera-ready). To verify the
results do not depend on this hardware choice, we additionally
re-trained the 1.2B baseline end-to-end on NVIDIA A100 GPUs
(8$\times$A100 80GB) under bit-identical model code, optimizer
recipe, tokenizer, data shuffle, and 23B-token budget; the two
runs match to $1\!\times\!10^{-4}$ \climb-avg
(Appendix~\ref{app:a100}).

\paragraph{Software versions.}
Training: PyTorch 2.x, Transformers 4.x, FlashAttention 2.x;
evaluation: lm-evaluation-harness v0.4.x. Specific patch
versions and Docker images will be released alongside the code at
the camera-ready stage.

\appendix
\section{Data Composition}
\label{app:data}

\paragraph{Corpus.}
All experiments train on publicly-available shards of the
CLIMB-Mix-400B pretraining corpus~\citep{diao2025climb}, a mixed-domain
English collection released alongside the CLIMB clustering-based
data-mixture method. CLIMB-Mix-400B is curated by domain-conditional
clustering of web, code, math, encyclopedia, and dialogue text. We
adopt the 12-task downstream evaluation protocol from the same work
(referred to as \climb-12 in this paper, §\ref{sec:method-eval}). The 1.2B experiments use a fixed
23.28B-token subset; the 3B experiments use a 60.04B-token
superset of the same shards. Both subsets are drawn from the same
pre-shuffled shard sequence, so the 3B corpus contains the 1.2B
corpus verbatim as a prefix.

\paragraph{Tokenizer.}
We tokenize with the 65{,}664-entry BPE tokenizer released with
\citet{karpathy2025nanochat}'s open-source small-LM reproduction
stack. The tokenizer is shared across baseline and all
modifications; no tokenizer change is part of the experimental
variable.

\paragraph{Packing.}
Raw documents are concatenated with document separators and split
into 1024-token sequences (sequence length matches training
context; cross-document attention is permitted within a packed
sequence, matching standard practice for Llama-2-style pretraining).
Packing produces 474 shards for the 1.2B subset
($22{,}738{,}794$ sequences, $23.28$B tokens, $1.6\!\times\!10^{-4}\%$
tokens discarded by pack-boundary trimming) and 1268 shards for
the 3B superset ($58{,}637{,}625$ sequences, $60.04$B tokens,
$1.1\!\times\!10^{-4}\%$ discarded).

\paragraph{Shuffle.}
All runs use an identical shuffle RNG seed at the sampler level, so
the $i$th training example seen by method $A$ and method $B$ at step
$i$ is the same token sequence. Per-seed variants
(baseline 43, 44; QK-Norm 43, 44; Softpick 43, 44) use the seed to
initialize the weight and dropout RNGs; the shuffle seed is held
fixed across all runs so that seed effects are isolated from
curriculum effects.

\paragraph{Validation.}
A held-out $131{,}072$-sequence validation pack from the same
corpus is used to compute validation loss at every 100-step
checkpoint. This is distinct from the downstream \climb{}-12
evaluation (§\ref{sec:method-eval}), which uses task-specific
public evaluation sets.

\paragraph{Reproducibility.}
The packing metadata (one JSON file each for 1.2B and 3B) is
released with the code. These include the shard count, total
tokens, tokenizer reference, sequence length, and the tokens
discarded by pack trimming, so a re-pack from the CLIMB-Mix-400B
public shards reproduces our exact training stream.

\section{Implementation Notes}
\label{app:impl}

This appendix describes, per method, how our implementation maps to
the published algorithm. This is especially important for methods
whose 1.2B \climb-avg departs from the original paper's claim, and
for methods that required light adaptation to fit the shared
training graph.

\subsection{AttnRes (pseudo-query variant)}
\label{app:impl-attnres}

The AttnRes implementation follows \citet{kimi2026attnres}
Equation~2 and Table~4 row~2 (``AttnRes Full, pseudo-query'').
At layer~$\ell$ the residual output is
\[
\alpha_{i\to \ell} =
  \text{softmax}_i\!\left(
    \frac{w_\ell^\top \cdot \text{RMSNorm}(v_i)}{\sqrt{d}}
  \right),
\]
\[
h_\ell = \sum_i \alpha_{i\to \ell}\, v_i,
\]
where $w_\ell \in \mathbb{R}^d$ is a learnable per-layer pseudo-query,
$v_i$ is the output of layer~$i$, and $i$ ranges over
$\{0,\dots,\ell-1\}$ plus the current layer's contribution.
Pseudo-query initialization: $w_\ell \sim \mathcal{N}(0, 0.02^2)$;
RMSNorm $\epsilon = 10^{-5}$.

\subsection{Softpick}

Softpick~\citep{zuhri2025softpick} is implemented as
$\text{softpick}(z) = \text{ReLU}(\text{softmax}(z) - 1/n)$, where
$n$ is the number of keys. No post-ReLU renormalization is
applied; this is the ``rectified softmax'' variant from the
original paper. All other attention plumbing (RoPE, GQA, KV cache)
matches the baseline full-attention implementation
line-by-line.

\subsection{QK-Norm}

QK-Norm applies an RMSNorm to $Q$ and $K$ independently, along the
head dimension, after the linear projection and before
RoPE~\citep{henry2020query, dehghani2023vit22b}. Norm parameters
are shared across heads within each projection and initialized to
unity. No learnable temperature is added.

\subsection{Selective Attention}

Selective Attention follows \citet{leviathan2024selective} with
one adaptation: instead of accumulating the selective mask~$S$
across layers (which would require modifying the shared
cross-layer interface), the mask is computed and applied
\emph{within} each layer. Heads are split into an attention group
and a masking group; masking-head scores are passed through ReLU,
zeroed on the diagonal and on the BOS column, and subtracted from
the attention-head logits. The Q/K projections are shared across
the two head groups, so the modification adds zero new parameters.

\subsection{Sigmoid Attention}

Sigmoid Attention replaces $\text{softmax}(QK^\top/\sqrt{d})$ with
$\text{sigmoid}(QK^\top/\sqrt{d} + b) / n$, where $b$ is a
learnable scalar initialized to zero and $n$ is the sequence
length~\citep{ramapuram2024sigmoid}. The $1/n$ normalization
keeps attention magnitudes comparable to softmax at matched
sequence lengths but does not enforce row sums to unity.

\subsection{Scalable-Softmax (SSMax)}

SSMax multiplies pre-softmax logits by $s \cdot \log n$, with
$s = \text{softplus}(s_\text{logit}) + 0.5$ a per-head learnable
scale~\citep{nakanishi2025ssmax}. $s_\text{logit}$ is initialized to
$0$, giving $s \approx 1.19$ at step 0. At the 1024-context
training length used here, $\log n \approx 6.9$, so the effective
temperature is $1 / (s \cdot \log n) \approx 0.12$ at
initialization, matching the $1/\sqrt{d_k}=0.125$ baseline
(per-head $d_k=64$ at hidden $2048$, 32 heads) within $4\%$.

\subsection{Differential Transformer}

Diff-Attn computes two softmax distributions per head by splitting
$Q$ and $K$ (each into halves of width $d_h/2$) while keeping $V$
at full width. The output is
$(\text{softmax}(Q_1 K_1^\top/\sqrt{d}) - \lambda \cdot
  \text{softmax}(Q_2 K_2^\top/\sqrt{d})) V$,
with $\lambda$ reparameterized per the original paper's depth
schedule
$\lambda_\text{init} = 0.8 - 0.6 \exp(-0.3(\ell-1))$~\citep{ye2024diff}.
Per-head GroupNorm is applied to the attention output.

\subsection{Value-Residual}

Value-Residual implements the ResFormer V-residual of
\citet{zhou2025value}:
$U_\ell = \text{softmax}(Q_\ell K_\ell^\top / \sqrt{d})
((1-\lambda)V_\ell + \lambda V_1)$,
where $V_1$ is read from the first layer's outputs and $\lambda$
is a learnable scalar initialized to $0.5$. Layer~0 degenerates
to standard attention because no first-layer reference is yet
available.

\subsection{DenseFormer}

DenseFormer replaces the standard residual with
$Y_\ell = \sum_{j=0}^{\ell} \alpha_{\ell,j} X_j$, where
$\{\alpha_{\ell,j}\}$ are learnable scalars initialized to the
identity ($\alpha_{\ell,\ell}=1$, others
$0$)~\citep{pagliardini2024denseformer}. This adds
$N(N+1)/2 = 300$ parameters at 24 layers and does not alter
per-step FLOPs.

\subsection{LayerScale}

LayerScale multiplies the sublayer output by a learnable
per-channel gate $\gamma$ initialized at $10^{-4}$, giving
$x_{\ell+1} = x_\ell + \gamma \odot f(x_\ell)$~\citep{touvron2021layerscale}.
$\gamma \in \mathbb{R}^d$ is a hidden-size-dimensional vector.
Initialization value matches the CaiT paper.

\subsection{HyperConnections}
\label{app:impl-hyper}

HyperConnections implements a lightweight output-side two-lane
variant of \citet{zhu2024hyper}. A fast lane follows the standard
residual; a slow lane maintains an exponential moving average of
sublayer outputs with learnable rate
$\beta = \sigma(\beta_\text{logit})$,
$\beta_\text{logit}$ initialized to $-2.2$
($\beta_0 \approx 0.1$). The slow lane contributes to the output
with weight $\alpha$ initialized to $0$, so the module reduces to
standard residual at step~0. The full-rank input-side lane
mixing~$A$ and output-side re-mixing $R$ from the original paper
are not reimplemented, because our shared residual interface does
not currently expose the sublayer input. The full diagnostic in
Appendix~\ref{app:diagnostics} discusses this together with other
reasons the 1.2B dense setting here differs from the paper's 671B
MoE setting.

\subsection{GeGLU}

GeGLU substitutes GELU for SiLU in the SwiGLU form:
$y = W_\text{down}(\text{GELU}(W_\text{gate} x) \odot W_\text{up} x)$~\citep{shazeer2020glu}.
The intermediate size matches the SwiGLU baseline (5632 at 1.2B);
parameter count and per-step FLOPs are identical.

\subsection{Softmax-Cap}

Softmax-Cap clips pre-softmax logits element-wise at $\pm 50$, a
recipe adopted in several long-context training reports to avoid
bf16 overflow. At the 1024-context used here, the cap is almost
never active; it is included as a minimal-surface-area baseline
variant.

\subsection{Sandwich Normalization}
\label{app:impl-sandwich}

Sandwich Norm follows \citet{ding2021cogview}: each sublayer
output is bracketed by two RMSNorms, one before the sublayer and
one before the residual add, giving
$x_{\ell+1} = x_\ell + \text{RMSNorm}(f(\text{RMSNorm}(x_\ell)))$.
This is distinct from HybridNorm in that both attention and FFN
use the same sandwich wrapper; it adds zero new parameters beyond
the extra norm scales.

\subsection{HybridNorm}
\label{app:impl-hybrid}

HybridNorm follows \citet{zhuo2025hybridnorm}: the attention block
uses Pre-LN ($\text{RMSNorm}$ before projection, residual add
after), while the FFN block uses Post-LN ($\text{RMSNorm}$ after
residual add). At 1.2B this configuration clears Bonferroni
significance ($z=+3.19$, $p_\text{Bonf}=0.027$); at 3B it diverges
at step 3500 with a 4.4$\times$ grad-norm spike, consistent with
the Post-LN deep-layer variance growth documented by
\citet{xiong2020layernorm} (§\ref{sec:3b-failures}).

\subsection{\texorpdfstring{ReLU$^{2}$ activation}{ReLU2 activation}}
\label{app:impl-primer}

ReLU$^{2}$ implements the non-gated two-matrix FFN from
\citet{so2022primer}: $y = W_\text{down}(\text{ReLU}(W_\text{up} x))^{2}$.
We match baseline parameter count and FLOPs by expanding the
intermediate dimension to $1.5 \times d_\text{inter}$, so that
$2 H d_\text{inter-primer} = 3 H d_\text{inter-SwiGLU}$. At 1.2B
it reaches $z=+2.36$ (nominally significant, does not clear
Bonferroni).

\subsection{Gated Attention with QK-Norm}
\label{app:impl-gated}

Gated + QK-Norm follows \citet{qiu2025gated}: a per-head sigmoid
gate multiplies the attention output before the output projection,
$h = (\sigma(W_g x)) \odot \text{Attn}(QKV)$, with $W_g$ applied
head-wise. QK-Norm is stacked on the Q/K projections as described
above. The combination adds approximately $0.1\%$ to both
parameter count and per-step FLOPs; at 1.2B it lands at $z=+1.23$,
indistinguishable from baseline under Bonferroni.

\subsection{Combination variants (Selective + QK-Norm; QK-Norm + GeGLU)}

The two combination variants reuse the components specified above
without modification. \emph{Selective + QK-Norm} stacks Selective
Attention's per-position learnable mask on top of the QK-Norm
RMSNorm of $Q$/$K$ projections; the mask is applied to the
post-softmax attention weights, after QK-Norm has stabilized the
pre-softmax logits. \emph{QK-Norm + GeGLU} replaces both the
attention's $Q$/$K$ normalization and the FFN gating in a single run,
matching the production stack used by Llama-3 and Gemma-2. Neither
variant introduces components not already documented in
§\ref{app:impl-gated} above; we list them separately because their
1.2B--3B ranking trajectories (rank 10$\to$3 and 7$\to$1
respectively) are central to the rank-reshuffle finding of
§\ref{sec:3b-top5}.

\section{FLOPs Accounting}
\label{app:flops}

\subsection{Per-step FLOPs derivation}

We decompose one training-step forward pass into attention-score,
attention-projection, and FFN components:

\begin{align*}
F_\text{attn-score}(\ell) &= 2 \cdot s^2 \cdot d \\
F_\text{attn-proj}(\ell)  &= 2\cdot(2 d^2 + 2 d \cdot d_\text{kv}) \\
F_\text{FFN}(\ell)        &= 2 \cdot k_\text{ffn} \cdot d \cdot d_\text{inter}
\end{align*}

where $s$ is sequence length, $d$ is hidden size, $d_\text{kv}$
accounts for grouped-query attention, $d_\text{inter}$ is FFN
intermediate size, and $k_\text{ffn}$ is 3 for three-matrix
variants (SwiGLU, GeGLU, QK-Norm + GeGLU) and 2 for the
two-matrix ReLU$^{2}$ with $1.5\times$ intermediate width
(yielding the same $k \cdot d_\text{inter}$ product).

The per-sequence forward FLOPs total to
$\sum_\ell [F_\text{attn-score}(\ell) + s\cdot(F_\text{attn-proj}(\ell) + F_\text{FFN}(\ell))]$,
times 3 for forward, backward, and activation-checkpointing
accounting.

Our computations are per-method and use each config's actual
$d$, $d_\text{inter}$, number of heads, and GQA ratio.

\subsection{Per-method FLOPs and parameter counts}

Table~\ref{tab:flops} reports per-step FLOPs and total parameters for
each of the 20 methods in the main comparison.

\begin{table*}[!t]
\centering
\small
\setlength{\tabcolsep}{6pt}
\begin{tabular}{@{}llrrr@{}}
\toprule
\textbf{Method} & \textbf{Category} & \textbf{Params} & \textbf{$\Delta$P} & \textbf{$\Delta$F} \\
\midrule
baseline            & ref        & 1.217\,B & $+0.00\%$   & $+0.00\%$ \\
softpick            & attention  & 1.217\,B & $+0.00\%$   & $+0.00\%$ \\
qknorm              & attention  & 1.217\,B & $+0.00\%$   & $+0.00\%$ \\
selective\_attn     & attention  & 1.192\,B & $-2.07\%$   & $+0.00\%$ \\
selective\_qknorm   & attention  & 1.192\,B & $-2.07\%$   & $+0.00\%$ \\
value\_residual     & attention  & 1.217\,B & $+0.00\%$   & $+0.00\%$ \\
diff\_attn          & attention  & 1.217\,B & $+0.01\%$   & $+0.00\%$ \\
sigmoid\_attn       & attention  & 1.217\,B & $+0.00\%$   & $+0.00\%$ \\
ssmax               & attention  & 1.217\,B & $+0.00\%$   & $+0.00\%$ \\
softmax\_cap        & attention  & 1.217\,B & $+0.00\%$   & $+0.00\%$ \\
gated\_attn\_qknorm & attention  & 1.218\,B & $+0.13\%$   & $+0.13\%$ \\
geglu\_ffn          & ffn        & 1.217\,B & $+0.00\%$   & $+0.00\%$ \\
qknorm\_geglu       & ffn        & 1.217\,B & $+0.00\%$   & $+0.00\%$ \\
relu\_squared & ffn      & 1.217\,B & $+0.00\%$   & $+0.00\%$ \\
sandwich\_norm      & norm       & 1.217\,B & $+0.01\%$   & $+0.00\%$ \\
hybrid\_norm        & norm       & 1.217\,B & $+0.00\%$   & $+0.00\%$ \\
denseformer         & residual   & 1.217\,B & $+0.00\%$   & $+0.00\%$ \\
layerscale          & residual   & 1.217\,B & $+0.00\%$   & $+0.00\%$ \\
hyper               & residual   & 1.217\,B & $+0.00\%$   & $+0.00\%$ \\
attnres             & residual   & 1.217\,B & $+0.02\%$   & $+0.00\%$ \\
\bottomrule
\end{tabular}
\caption{Per-method parameter count and per-step training FLOPs
relative to baseline, for the 20 methods in the main comparison.
Columns: method, taxonomy category, parameter count, $\Delta$P
(\% vs baseline params), $\Delta$F (\% vs baseline per-step FLOPs).
All 20 methods are within $\pm 0.13\%$ of the baseline on both
parameters and per-step FLOPs; Selective is $2.07\%$ cheaper by
parameter count.}
\label{tab:flops}
\end{table*}

\section{Cross-Hardware Baseline Reproduction}
\label{app:a100}

To test whether our results carry hardware-specific numerical
sensitivity, we re-trained the 1.2B baseline end-to-end on NVIDIA
A100 GPUs under the bit-identical model code, optimizer recipe,
tokenizer, data shuffle, and 23B-token budget used for the primary
accelerator. The A100 run uses the same Llama-style 1.2B
configuration described in §\ref{sec:method-training} (24 layers,
$d=2048$, 16 heads, 4 KV groups, RoPE, GQA, SwiGLU, RMSNorm), the
same AdamW recipe ($\beta_1{=}0.9$, $\beta_2{=}0.95$, weight
decay $0.1$), the same cosine learning-rate schedule with linear
warm-up, and the same $44{,}000$-step training horizon. The only
intentional differences are the accelerator vendor (NVIDIA A100
vs.\ the primary accelerator) and the precision pathway
($\textrm{bf16}$ tensor cores on both, but with vendor-specific
fused kernels).

\begin{table}[!t]
\centering
\small
\setlength{\tabcolsep}{4pt}
\begin{tabular}{@{}lrrr@{}}
\toprule
\textbf{Task} & \textbf{Primary} & \textbf{A100} & \textbf{$\Delta$} \\
\midrule
piqa            & 0.7492 & 0.7448 & $-0.0044$ \\
arc\_challenge  & 0.3933 & 0.3933 & $\phantom{-}0.0000$ \\
arc\_easy       & 0.7226 & 0.7243 & $+0.0017$ \\
hellaswag       & 0.5589 & 0.5564 & $-0.0025$ \\
winogrande      & 0.5620 & 0.5406 & $-0.0213$ \\
social\_iqa     & 0.4176 & 0.4130 & $-0.0046$ \\
mmlu            & 0.2736 & 0.2685 & $-0.0051$ \\
openbookqa      & 0.3840 & 0.3720 & $-0.0120$ \\
boolq           & 0.5872 & 0.6165 & $+0.0294$ \\
race            & 0.3282 & 0.3435 & $+0.0153$ \\
lambada\_openai & 0.4174 & 0.4203 & $+0.0029$ \\
truthfulqa\_mc2 & 0.3905 & 0.3919 & $+0.0014$ \\
\midrule
\textbf{\climb-avg} & \textbf{0.4820} & \textbf{0.4821} & \textbf{$+0.0001$} \\
\bottomrule
\end{tabular}
\caption{1.2B baseline cross-hardware reproduction. ``Primary''
denotes the AI accelerator used in the main 1.2B and 3B
experiments (vendor anonymized for review); ``A100'' is an end-to-end
re-training on NVIDIA A100 GPUs under bit-identical model code,
optimizer recipe, tokenizer, data shuffle, and 23B-token budget.
\climb-avg differs by $1{\times}10^{-4}$ between the two runs, and
per-task deltas span $\pm 0.029$ with no systematic direction.}
\label{tab:a100}
\end{table}

Table~\ref{tab:a100} reports the per-task \climb{} accuracies on
both platforms. The aggregate \climb-avg differs by $1{\times}10^{-4}$
($0.4820$ vs.\ $0.4821$), within the third-decimal precision of
the metric. Per-task deltas span $\pm 0.029$ with no systematic
direction: the primary platform is higher on PIQA, HellaSwag,
WinoGrande, SocialIQA, MMLU, and OpenBookQA; A100 is higher on
ARC-E, BoolQ, RACE, LAMBADA, and TruthfulQA-MC2; ARC-C is
identical.

A separate validation-loss check yields $\Delta_{\text{val-loss}}
= 0.015$ between the two runs, with the primary platform marginally
lower. This loss gap does not propagate to the downstream metric
in the same direction or magnitude---a pattern we already document
in Appendix~\ref{app:diagnostics} for several attention-output
modifications, and which here serves as an additional null
control: the loss difference between two baseline trainings on
different hardware is smaller than several of the loss--downstream
gaps we report on attention-output methods.

This cross-hardware reproduction is the strongest single piece of
evidence against an infrastructure-artefact explanation for the
three 3B divergences (HybridNorm, SSMax,
HyperConnections, §\ref{sec:3b-failures}). The baseline reproduces
to within $10^{-4}$ on A100 under the same recipe, while the
divergences each have documented theoretical fragility (Post-LN
variance growth at depth~\citep{xiong2020layernorm},
softmax-replacement logit blowup~\citep{ramapuram2024sigmoid},
multi-channel EMA drift~\citep{zhu2024hyper}) that is independent
of accelerator vendor. We did not re-run the modified methods at
1.2B on A100 under the same protocol: the cost of replicating a
20-method suite on a second hardware brand is comparable to the
original 1.2B run. The single-method (baseline) cross-hardware
check is intended as a sanity bound on the platform-as-confound
hypothesis, not as a full replication.


\section{Diagnostic Analysis of Significant Failures}
\label{app:diagnostics}

\begin{table*}[!t]
\centering
\small
\setlength{\tabcolsep}{5pt}
\begin{tabular}{@{}llrll@{}}
\toprule
\textbf{Method} & \textbf{Seed} & \textbf{NaN step} & \textbf{Pre-NaN grad signature} & \textbf{Mechanism} \\
\midrule
HybridNorm & 42 & 3{,}500  & $0.19 \to 0.83 \to$ NaN (1 step)      & \multirow{3}{*}{Post-LN FFN variance growth} \\
HybridNorm & 43 & 2{,}900  & $0.19 \to$ NaN (no intermediate spike)  & \\
HybridNorm & 44 & 2{,}800  & $0.18 \to$ NaN, then sustained $\sim$14 & \\
\midrule
SSMax        & 42 & 5{,}800  & monotone rise $27 \to 1578$ over 1{,}800 steps & Softmax-replacement logit blowup \\
HyperConnections        & 42 & 18{,}300 & sustained $>100$ peaks from step 15{,}000 & Multi-lane EMA residual drift \\
\bottomrule
\end{tabular}
\caption{3B training divergences under the shared iso-recipe.
HybridNorm is reported across all three baseline
noise-floor seeds (42, 43, 44); all
three diverge in a 2800--3500 step window with pre-NaN
grad-norm in the $0.18$--$0.19$ range, identical to baseline's
stable trajectory. Seed 42
shows a single-step spike visible at 100-step logging; seeds 43
and 44 collapse directly from the stable range without a visible
intermediate step. SSMax and HyperConnections are each
reported on the single seed that was attempted, with the pre-NaN
grad-norm signatures listed above.}
\label{tab:3b-failures}
\end{table*}

\begin{figure}[!t]
  \centering
  \includegraphics[width=\columnwidth]{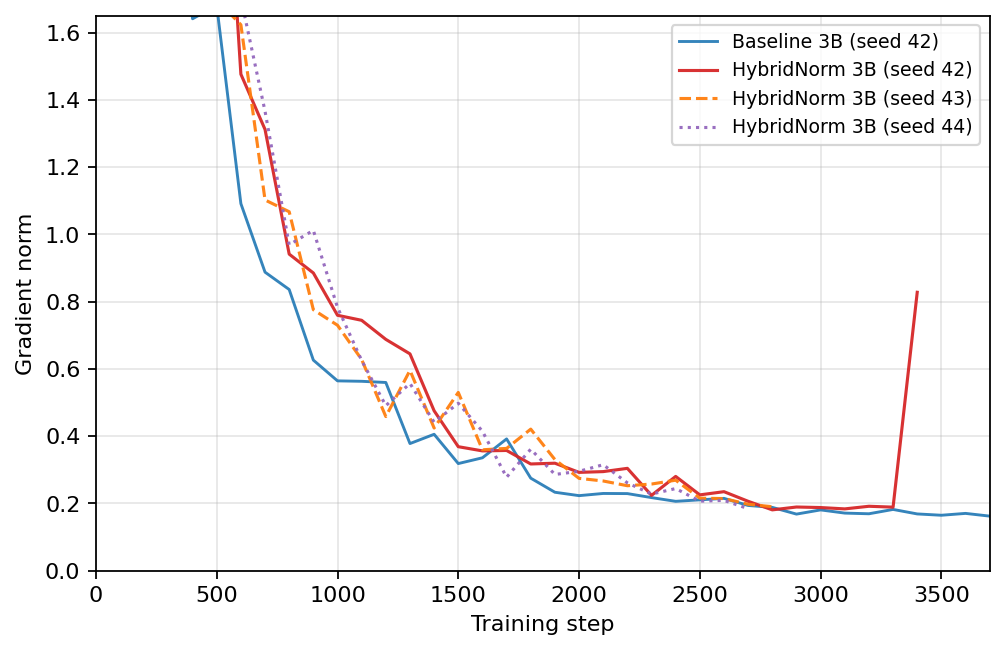}
  \caption{3B training grad-norm through step 3700 for the three
  baseline noise-floor seeds. Baseline (blue) tracks a stable
  $\sim$$0.15$--$0.20$ range. HybridNorm under seeds 42 (red,
  solid), 43 (orange, dashed), and 44 (violet, dotted) each track
  the same range before collapsing to NaN; each terminal
  $\times$ marks the last finite step before divergence at
  steps 3500, 2900, and 2800 respectively (seed~42 first
  spikes to grad-norm $0.83$ at step 3400 before NaN at 3500).
  All three diverge shortly after the cosine peak. Pre-NaN
  grad-norm $0.18$--$0.19$ across seeds and a narrow divergence
  window indicate Post-LN-on-FFN variance growth at 28-layer 3B
  depth~\citep{xiong2020layernorm} rather than seed noise.}
  \label{fig:hybrid-spike}
\end{figure}

Six of the 20 modifications we tested significantly underperform
baseline at 1.2B (Table~\ref{tab:main-results}). Two are softmax
replacements (Sigmoid Attention, SSMax); one is a
softmax-subtraction variant (Diff-Attn); and three are
residual-connection modifications (AttnRes,
LayerScale, HyperConnections). This section examines
\emph{how} these failures manifest. Two views are used: final
validation loss compared to \climb{}-avg
(Figure~\ref{fig:loss-curves}), and the per-task delta heatmap
(Figure~\ref{fig:heatmap}).

\subsection{Validation loss does not predict the attention-side failures}
\label{sec:diag-loss}

\begin{figure}[!t]
  \centering
  \includegraphics[width=\columnwidth]{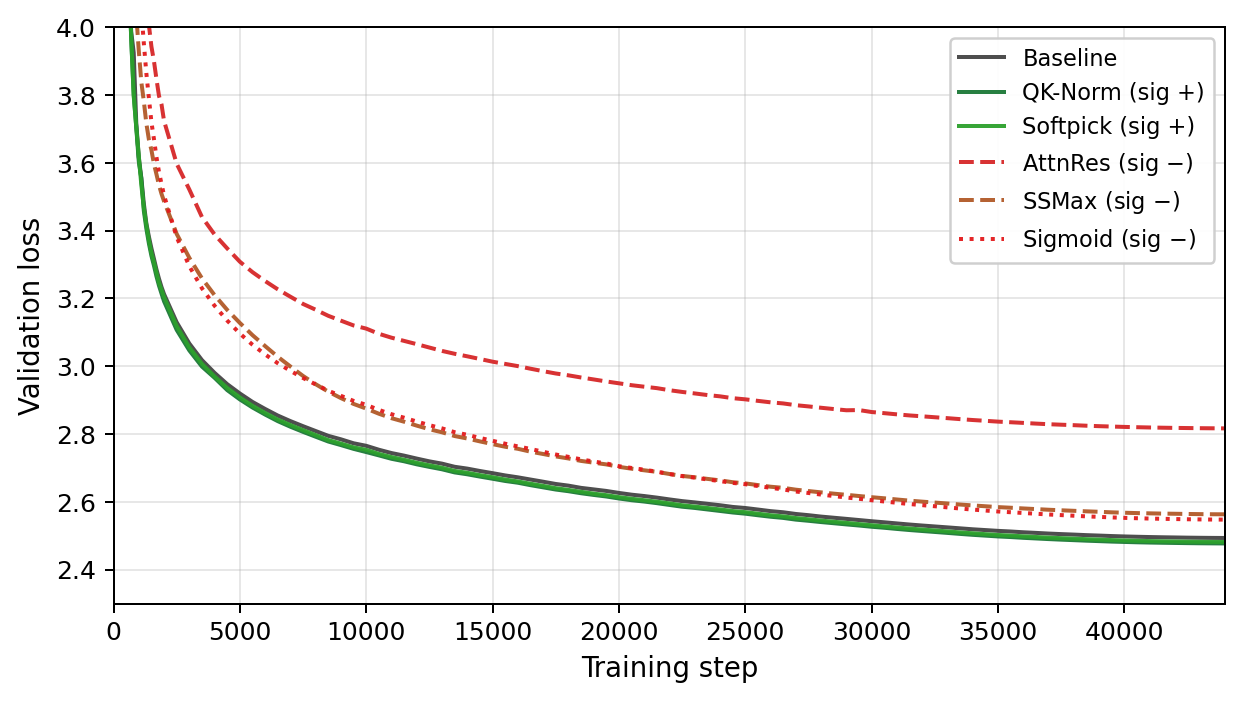}
  \caption{Validation-loss curves for six representative methods at
  1.2B over 44{,}000 training steps. The two attention-side failures
  (SSMax, Sigmoid Attention) converge to within $\sim$0.07 nats of
  baseline---a 2--3\% relative gap. The residual-side failure
  (AttnRes) ends 0.33 nats above baseline. The two improvers
  (QK-Norm, Softpick) track baseline within visual resolution.}
  \label{fig:loss-curves}
\end{figure}

Figure~\ref{fig:loss-curves} plots validation loss against training
step for six methods: two significant winners (QK-Norm, Softpick),
baseline, and three significant failures. At the end of training
the final losses are close for the two attention-side failures but
not for AttnRes (Table~\ref{tab:loss-vs-climb}).

\begin{table}[!t]
\centering
\small
\begin{tabular}{@{}lrrr@{}}
\toprule
\textbf{Method} & \textbf{val loss} & \textbf{\climb-avg} & \textbf{$z$} \\
\midrule
Softpick          & 2.4814 & 0.4922 & $+4.47$ \\
QK-Norm           & 2.4771 & 0.4882 & $+2.51$ \\
Baseline          & 2.4890 & 0.4829 & \hspace{1.2em}$0.00$ \\
Sigmoid Attention & 2.5474 & 0.3217 & $-77.7$ \\
SSMax             & 2.5630 & 0.4208 & $-29.9$ \\
AttnRes           & 2.8165 & 0.4218 & $-29.5$ \\
\bottomrule
\end{tabular}
\caption{Final validation loss, \climb-avg, and bootstrap $z$-score
for six representative methods at 1.2B.}
\label{tab:loss-vs-climb}
\end{table}

Sigmoid Attention's validation loss exceeds baseline's by only
2.4\%, and SSMax's by 3.0\%. A pretraining-perplexity ranking---the
primary metric in \citet{narang2021transformer}---would place both
methods near baseline. Their \climb{}-avg scores, however, drop by
16 and 6 points respectively, on a scale where the entire 1.2B range
across our 20 methods is roughly three points.

The loss-downstream decoupling observed here is therefore specific
to attention-output modifications: Sigmoid Attention and
SSMax reach near-baseline losses while failing downstream.
The third failure in the figure, AttnRes, behaves
differently. Its validation loss is $0.33$ nats above baseline (13\%
relative) and its \climb-avg drop tracks that loss gap; a
loss-based ranking would correctly flag AttnRes as
substantially worse than baseline. The decoupling claim is not a
universal loss-is-unreliable statement---it is a scoped observation
that methods that replace or post-process the attention-output
distribution can reach low loss without transferring that low loss
to downstream accuracy. This narrows but does not contradict
\citet{tay2023scaling}, who document a smaller-magnitude version of
the same gap across ten architecture families (15M to 40B
parameters). The magnitude observed here (up to 16 \climb-points)
is several times the gap they document.

\subsection{Failure signatures}
\label{sec:diag-signatures}

\begin{figure*}[!t]
  \centering
  \includegraphics[width=\textwidth]{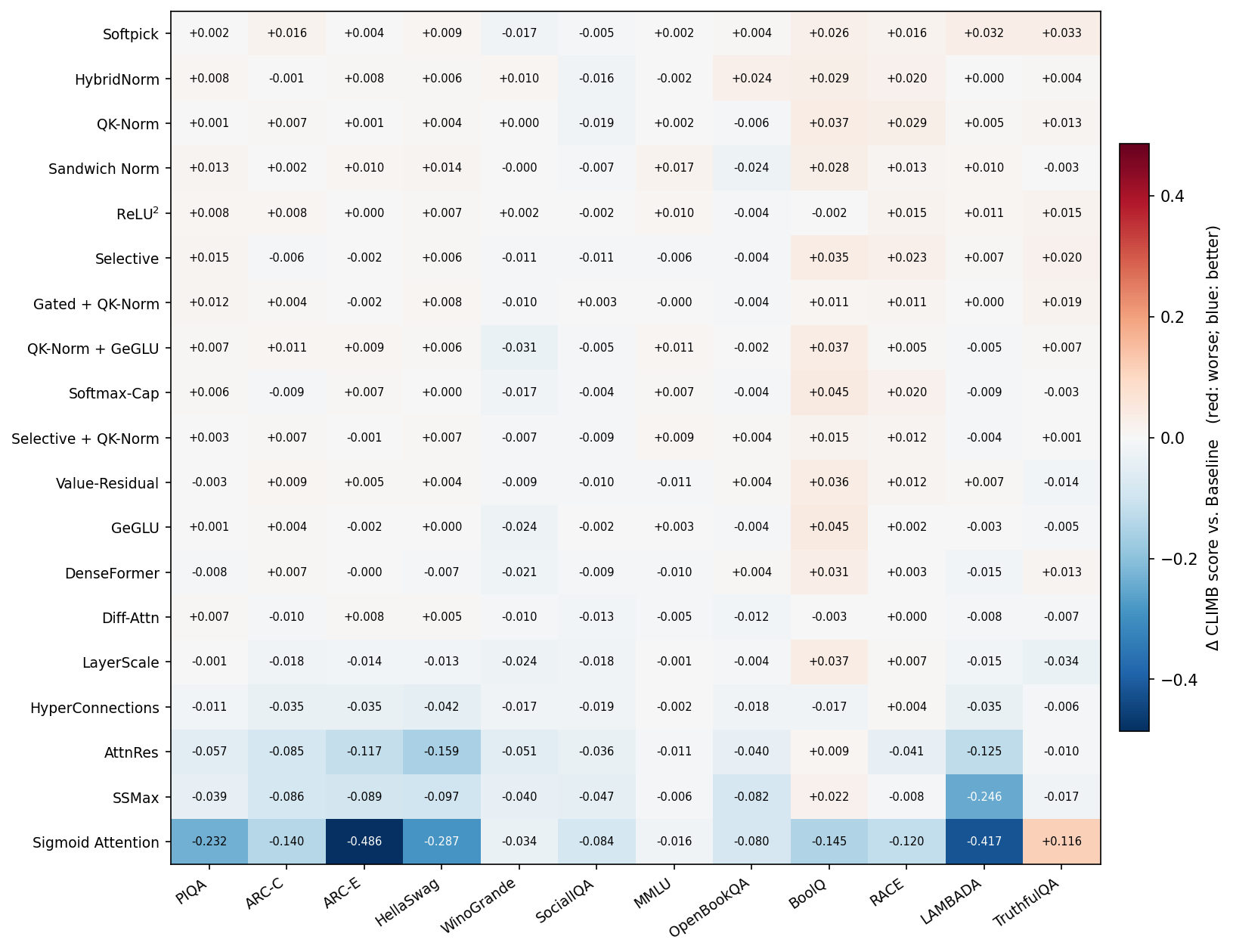}
  \caption{Per-task \climb{} deltas vs baseline at 1.2B. Rows are
  methods sorted by \climb-avg; columns are the 12 \climb{} tasks.
  Red cells indicate accuracy below baseline, blue above.}
  \label{fig:heatmap}
\end{figure*}

\paragraph{Sigmoid Attention: LAMBADA drop.}
The accuracy drop for Sigmoid Attention is not uniform across tasks
(Figure~\ref{fig:heatmap}). LAMBADA (OpenAI) falls from baseline's
$0.42$ to near $0$; the other 11 tasks each lose $0.03$--$0.08$
points. LAMBADA requires predicting the final word of a passage
given its preceding narrative, which requires attention to
concentrate on a specific retrieval-relevant token. Without softmax's
normalization, Sigmoid Attention cannot guarantee that attention
weights sum to one, and in practice produces more diffuse attention
distributions that fail to select a single key position. General
language modelling behaves adequately---the conditional distribution
over next tokens is still usable for most \climb{} tasks---but
this specific kind of look-up fails.

\paragraph{SSMax: uniform degradation.}
SSMax's drops are smaller but uniform: $0.014$--$0.025$ points on
nearly every task (Figure~\ref{fig:heatmap}, row SSMax).
SSMax replaces softmax with a variant that includes an additional
learned per-head temperature intended to prevent entropy collapse
at long contexts~\citep{nakanishi2025ssmax}. In our 1024-token-context,
1.2B setting the temperature does not stabilize near softmax's
effective value; the result is a small but consistent drift of
attention away from the per-task optimum. No single task shows a
localized drop, but the aggregate drop is large.

\paragraph{AttnRes: broad regression, not a local bug.}
AttnRes~\citep{kimi2026attnres} replaces the identity
residual after attention with a learned cross-layer mixture,
weighted by a pseudo-query dot-producted against previous layers'
outputs. We verified our implementation matches the pseudo-query
variant in Equation~2 and Table~4 row 2 of the original paper
line-by-line (Appendix~\ref{app:impl-attnres}); the $z=-29.5$ result
is not attributable to a coding error. Unlike Sigmoid Attention and
SSMax, AttnRes's validation loss also rises substantially
(0.33 nats higher than baseline), and its \climb{} drops track the
loss gap. The model is globally worse at next-token prediction, not
subtly misbehaving on a specific downstream skill.
\citet{kimi2026attnres} report gains at larger MoE scale on GPQA;
our result scopes that claim: at 1.2B dense on \climb{}-12,
pseudo-query cross-layer residual does not help.

\paragraph{Diff-Attn: subtraction of two softmaxes.}
Differential attention~\citep{ye2024diff} computes two softmax
distributions per head and subtracts one from the other. Each
individual softmax is preserved, but the subtraction produces
un-normalized attention weights that need not sum to one and can be
negative; under the operational criterion of
§\ref{sec:method-taxonomy} it is therefore \emph{hard}. The 1.2B
result ($z=-2.45$) is significantly below baseline. One
plausible mechanism is that when the two subtracted heads become
correlated during training, the subtraction amplifies noise rather
than cancelling distractors, but the single-seed experiment reported
here cannot distinguish this from alternative explanations.

\paragraph{LayerScale: uniform degradation, mechanism unresolved.}
LayerScale~\citep{touvron2021layerscale} multiplies the sublayer
output by a learnable per-channel gate $\gamma$ initialized at
$10^{-4}$, giving $x_{\ell+1} = x_\ell + \gamma \odot f(x_\ell)$.
The small-init design intends to let deep networks train stably by
starting near the identity and learning per-layer contribution
magnitudes. In the 1.2B setting the $z=-4.42$ degradation is
accompanied by a roughly uniform $0.01$--$0.02$ drop across
\climb{} tasks (Figure~\ref{fig:heatmap}), rather than a
task-localized failure. The failure signature is therefore
``global under-performance'' rather than a specific downstream skill
loss. One possible mechanism is that under a fixed optimizer recipe
tuned for the identity residual, the small-init residual branch
trains more slowly and does not close the gap within 23B tokens;
CaiT reports gains at hyperparameters co-tuned with the
LayerScale init value, which is not varied here. Distinguishing
this hypothesis from alternatives (effective-depth change,
initialization interacting with weight decay, etc.) would require
an init sweep that is left to future work.

\paragraph{HyperConnections: multi-lane overhead at dense 1.2B.}
HyperConnections~\citep{zhu2024hyper} replaces the single residual
stream with $n$ parallel lanes at different update rates, mixed at
both the sublayer input (matrix $A$) and output (matrices $R$, $D$).
The original paper reports gains on a 671B-parameter MoE and shows
increasing benefit at larger scale. Our implementation uses the
$n{=}2$ variant with slow-lane blend rate $\beta$ and output weight
$\alpha$ initialized so that the module reduces to standard residual
at step 0. Despite that identity-preserving initialization, the
1.2B dense run degrades uniformly across \climb{} tasks with
$z=-9.79$ (Figure~\ref{fig:heatmap}, row HyperConnections). Three
conditions present in \citet{zhu2024hyper}'s 671B setting are absent
from ours: (i) mixture-of-experts routing, which changes what a
residual stream needs to carry; (ii) model depth of $\sim$60 layers,
where the lane mechanism has more room to differentiate; (iii)
separate warmup for the lane-mixing parameters. The result here is
consistent with a scale-dependent benefit that 24-layer dense 1.2B
does not capture, rather than a failure of the method in general.

\subsection{A common thread: sharp attention and matched recipe}
\label{sec:diag-principle}

Two patterns emerge across the six failure signatures above. The
three attention-side failures (Sigmoid Attention,
SSMax, Diff-Attn) each reduce softmax's ability
to concentrate attention on a single key position, and the
\climb{} tasks that require such concentration---LAMBADA and, to a
lesser degree, ARC-Easy and HellaSwag---show the largest per-method
drops in Figure~\ref{fig:heatmap}. The three methods that
significantly improve over baseline (QK-Norm, Selective Attention,
Softpick) all preserve this concentration ability while adding
regularization or filtering. The pattern is not airtight:
Diff-Attn preserves individual softmaxes but still fails,
suggesting that preserving the softmax computation is necessary but
not sufficient to preserve the output attention distribution's
useful properties.

The three residual-side failures (AttnRes,
LayerScale, HyperConnections) share a different signature:
each is reported to improve a \emph{different} base architecture
or scale---671B MoE, larger-MoE GPQA, co-tuned CaiT---and each
degrades uniformly rather than task-specifically when transplanted
into 1.2B dense under a fixed recipe. This is the scale-and-recipe
sensitivity documented by \citet{narang2021transformer}: a
modification that helps under its authors' setup does not
automatically help under a different scale and an unmodified
optimizer recipe.

\subsection{Per-task statistical cross-checks}
\label{sec:diag-stouffer}

The bootstrap $z$-scores in Table~\ref{tab:main-results} aggregate
across the 12 \climb-12 tasks via the macro-average. As a
complementary cross-check that does not depend on the macro-average
or the bootstrap reference distribution, we compute per-task
Welch's $t$-tests against the three baseline seeds and combine the
12 per-task one-sided $p$-values via Stouffer's
$Z\!=\!\sum_i \Phi^{-1}(1\!-\!p_i)/\sqrt{12}$, treating each task as
an independent piece of evidence. For QK-Norm this gives
$Z{=}{+}2.68$, one-sided $p{=}0.0037$ (two-sided $0.0074$); 9 of 12
tasks have higher QK-Norm mean than baseline (the three exceptions:
PIQA, ARC-C, SocialIQA). For Softpick the corresponding combination
is $Z{=}{+}2.85$, one-sided $p{=}0.0022$ (two-sided $0.0044$), with
11 of 12 task means above baseline (only SocialIQA below). Stouffer
combination is conservative under positively-correlated tasks, so
the $p$-values are upper bounds on the true significance. Both
methods remain significant under this alternative aggregation,
providing independent evidence beyond the macro-average bootstrap
and the 3-seed Welch's $t$ on \climb-avg reported in
§\ref{sec:method-noise}.

\end{document}